\documentclass{article}

\usepackage[preprint]{neurips_2025}

\usepackage[utf8]{inputenc} 
\usepackage[T1]{fontenc}    
\usepackage{hyperref}       
\usepackage{url}            
\usepackage{booktabs}       
\usepackage{amsfonts}       
\usepackage{graphicx}       
\usepackage{nicefrac}       
\usepackage{microtype}      
\usepackage{xcolor}         
\usepackage{amsmath}
\usepackage{amssymb}
\usepackage{algorithm}  
\usepackage[noend]{algpseudocode}  
\usepackage{algorithmicx}
\usepackage{array}
\usepackage{colortbl}
\usepackage{natbib}
\usepackage{caption}

\newcommand{\systemname}{\text{TrajHF}}

\title{Learning Personalized Driving Styles via Reinforcement Learning from Human Feedback}

\author{%
  Derun Li\textsuperscript{1,2\thanks{Equal contribution.}}\And
  Changye Li\textsuperscript{3\footnotemark[1]}\And
  Yue Wang\textsuperscript{4\footnotemark[1]}\And
  Jianwei Ren\textsuperscript{2}\And
  Xin Wen\textsuperscript{4}\And
  Pengxiang Li\textsuperscript{4,5}\And
  Leimeng Xu\textsuperscript{4}\And
  Kun Zhan\textsuperscript{4}\And 
  Peng Jia\textsuperscript{4}\And  
  Xianpeng Lang\textsuperscript{4}\And
  Ningyi Xu\textsuperscript{1}\And
  Hang Zhao\textsuperscript{2,5\thanks{Corresponding Author. \texttt{hangzhao@mail.tsinghua.edu.cn}}}\AND
  \\
  \textsuperscript{1}Shanghai Jiao Tong University\quad
  \textsuperscript{2}Shanghai Qi Zhi Institute\quad
  \textsuperscript{3}Peking University \\
  \textsuperscript{4}LiAuto\quad
  \textsuperscript{5}Tsinghua University
}

\begin{document}

\maketitle

\begin{abstract}
  Generating human-like and adaptive trajectories is essential for autonomous driving in dynamic environments. While generative models have shown promise in synthesizing feasible trajectories, they often fail to capture the nuanced variability of personalized driving styles due to dataset biases and distributional shifts. To address this, we introduce \systemname, a human feedback-driven finetuning framework for generative trajectory models, designed to align motion planning with diverse driving styles. \systemname\ incorporates multi-conditional denoiser and reinforcement learning with human feedback to refine multi-modal trajectory generation beyond conventional imitation learning. This enables better alignment with human driving preferences while maintaining safety and feasibility constraints. \systemname\ achieves performance comparable to the state-of-the-art on NavSim benchmark. \systemname\ sets a new paradigm for personalized and adaptable trajectory generation in autonomous driving.
\end{abstract}

\section{Introduction}

In autonomous driving, trajectory planning is responsible for generating safe, feasible, and human-like motions in complex and dynamic traffic environments. Heuristics-based learning methods~\citep{zhao2021tnt, gu2021densetnt,xing2025goalflow} perform well in structured scenarios but struggle to generalize across diverse driving conditions. 
Imitation learning, most prominently generative models~\citep{jiang2023motiondiffuser,jiang2024scenediffuser,ngiam2021scene,sun2023large}, provides a data-driven alternative. They synthesize diverse human-like trajectories, but do not explicitly capture the nuanced preferences that characterize human driving behaviors, such as individual tendencies, regional norms, and environmental factors.

\begin{figure}
    \centering
    \includegraphics[width=\linewidth]{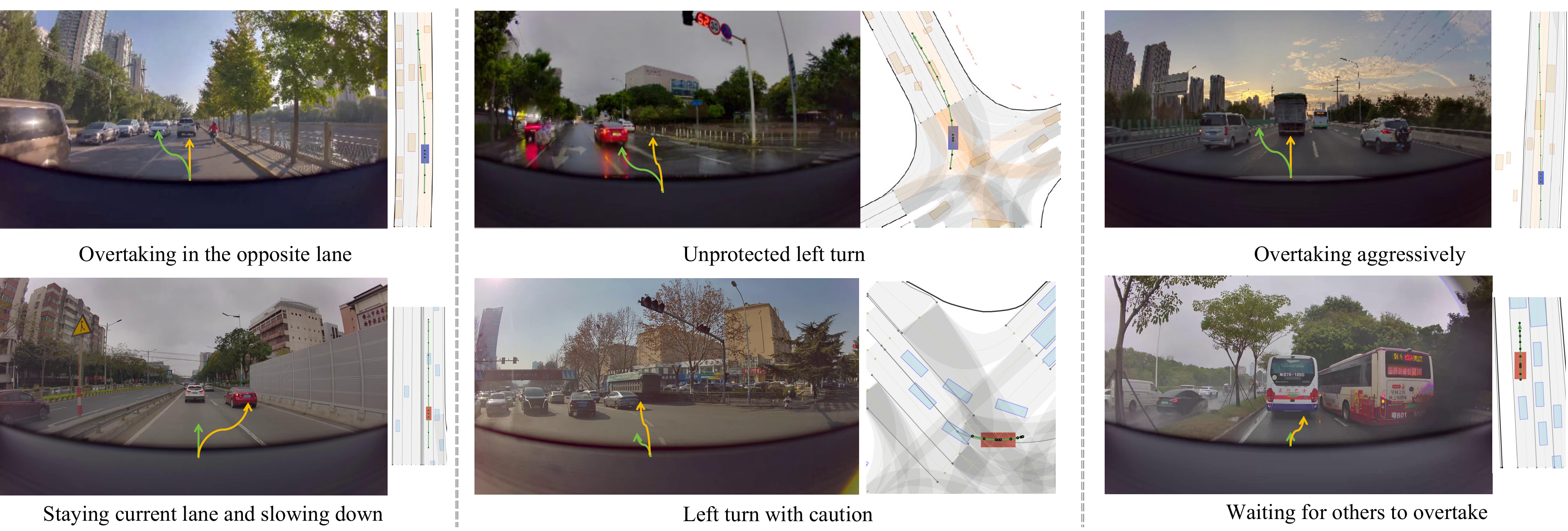}
    \caption{\textbf{Illustration of human preference problem in autonomous driving problem.} The first line is aggressive driving style examples and the second line is defensive style. Yellow lines indicate ``normal'' planning trajectories and green lines represent the stylized (``aggressive'' or ``defensive'') human preferred trajectories.}
    \label{fig1}
    \vspace{-1em}
\end{figure}

Learning from demonstrations is fundamentally constrained by dataset bias. Behavior cloning only captures an average driving behavior, failing to represent the full spectrum of maneuvers. Generative imitation learning, on the other hand, learns real data distributions in the dataset, but is usually dominated by frequent modes, and hardly performs the best behavior unless sampling is conducted in an exhaustive manner. 
This misalignment degrades model performance in scenarios requiring deviations from frequent demonstrations, such as aggressive adaptations in complex traffic interactions, as shown in Fig.~\ref{fig1}.  
Furthermore, human driving styles are influenced not only by kinematic constraints but also by higher-level cognitive and social factors, such as risk tolerance, interactions with other road users, and situational awareness. Existing methods struggle to encode these latent factors, leading to planned motions that, while technically feasible, may feel unnatural, overly conservative, or unpredictably assertive. This further reduces user trust and acceptability, underscoring the need for trajectory generation methods that align with diverse human driving behaviors.

To address this challenge, we propose \systemname, a novel framework to finetune generative trajectory models using human feedback to align trajectory generation with diverse personalized driving styles. Human feedback, in the form of comparative trajectory rankings or explicit annotations, provides a rich supervisory signal that extends beyond conventional imitation learning paradigms. By integrating this feedback into the finetuning process, our approach enables generative models to capture human-preferred driving styles while maintaining safety and feasibility constraints.

We systematically investigate methodologies for incorporating human feedback into trajectory model finetuning. We propose a novel Multi-Conditioned Denoiser (MDC) Transformer network to generate multi-modal trajectory planning. Our design of conditioning multi-modal information gains significant improvement of trajectory generation quality, reaching performance comparable to the state-of-the-art on NavSim~\citep{dauner2024navsim} benchmark with uniform sampling and optimization, without any explicit guidance like anchors.  Meanwhile, we explore reinforcement learning with human feedback and leverage group computation gathering rewards from each mode of generated trajectories to iteratively refine the generative trajectory distribution. Experimental results demonstrate that our method improves alignment with distinct driving styles in human preference data (``aggressive'' or ``defensive'', respectively) while preserving critical motion planning constraints. Using human feedback as an auxiliary learning signal, \systemname\ offers a more personalized and adaptable trajectory generation paradigm, paving the way for autonomous vehicles that better reflect human driving behavior while ensuring safety and efficiency.

The primary contributions of this paper are as follows:
\begin{enumerate}
    \item We systematically investigate the trajectory distribution shift problem between the average behaviors and the personalized driving styles.
    \item We propose a human feedback-driven finetuning framework for generative trajectory models, enabling alignment with diverse human driving preferences.
    \item Comprehensive experimental evaluations demonstrate the effectiveness of \systemname\ in generating human-aligned autonomous driving trajectories across diverse scenarios.
\end{enumerate}

\section{Related Work}

\subsection{End-to-end Trajectory Planning}
Imitation learning (IL) has been widely adopted for end-to-end trajectory planning from expert demonstrations~\citep{codevilla2018end}. UniAD~\citep{hu2023planning} and ViP3D~\citep{gu2023vip3d} leverage BEV inputs and transformer architectures to build planning-centric frameworks. VAD~\citep{jiang2023vad} and VADv2~\citep{chen2024vadv2} discretize trajectory space to transform regression into classification via trajectory vocabularies. NavSim~\citep{dauner2024navsim} introduces a benchmark focused on interaction-rich driving scenarios. Transfuser~\citep{chitta2022transfuser} fuses lidar and image features, and FusionAD~\citep{ye2023fusionad} propagates this fusion to downstream modules. Hydra-MDP~\citep{li2024hydra} distills multi-source knowledge through trajectory vocabulary. In contrast, our diffusion-based planner directly generates continuous, multi-modal trajectories without anchors or discretized spaces, supporting robust planning in complex scenes.

\subsection{Generative Trajectory Models}
Generative models provide a principled way to capture multi-modal driving behaviors. Earlier works use GANs~\citep{gupta2018social,fang2022atten} or VAEs~\citep{xu2022socialvae,de2022vehicles}, while recent advances rely on autoregressive~\citep{sun2023large} and diffusion models~\citep{jiang2023motiondiffuser,jiang2024scenediffuser}. In end-to-end settings, VBD~\citep{huang2024versatile} combines denoising with behavior prediction, DiffusionDrive~\citep{liao2024diffusiondrive} reduces computation via truncated inference, and GoalFlow~\citep{xing2025goalflow} integrates goal-conditioned flow matching. However, these models prioritize feasibility over personalization, lacking mechanisms to capture the diversity of human driving preferences, which is addressed in our proposed framework \systemname.

\subsection{Finetuning with Human Feedback}
Reinforcement Learning from Human Feedback (RLHF)~\citep{christiano2017deep} integrates subjective preferences via reinforcement learning and has shown success across domains~\citep{ouyang2022training,yu2024rlhf,zhang2024grape,shao2024deepseekmath}. Instead of relying on dense annotations, it optimizes models toward human-preferred distributions from limited feedback. Recent work applies RLHF to diffusion models~\citep{wallace2024diffusion} using preference optimization without explicit rewards. In driving tasks,\citet{wang2024reinforcement} train reward models on trajectory preferences, and\citet{Sun_2024} use coarse-grained safety preference data with KL penalties. We adopt GRPO~\citep{deepseek}, a critic-free method for efficient reward-based finetuning, and combine it with behavior cloning~\citep{ross2010efficient} to retain model capabilities.

\section{Problem Definition}

Let the state space of an autonomous vehicle be denoted as $S$, where each state $s_l\in S$ indicates the kinematic properties (e.g., position and heading) of the agent at timestamp $l$. A trajectory $x$ is defined as a sequence of states, 
\begin{equation}
\label{eq:traj}
    x = \{s_1, s_2, \dots, s_L\}
\end{equation}
where $L$ is the trajectory length. The expert trajectory distribution $P_{\mathrm{data}}(x)$ reflects human driving behaviors in the training dataset, whereas $P_{\theta}(x)$ represents the distribution induced by a generative trajectory model parameterized by $\theta$. In a typical imitation learning setup, the goal is to minimize the discrepancy between these distributions, which can be formalized as the Kullback-Leibler (KL) divergence:
\begin{equation}
\label{eq:KL}
    D_{KL}(P_{\mathrm{data}}\|P_\theta) = \sum_{x}P_{\mathrm{data}}(x)\log\frac{P_{\mathrm{data}}(x)}{P_{\theta}(x)}
\end{equation}

We believe that personalized driving styles can be depicted by human preferences when making decisions in complex driving scenarios. In this work, we deem styles and preferences as two equivalent concepts. Human driving preferences can be characterized by a latent distribution $P_{\mathrm{pref}}(x)$, with each driver or style corresponding to a distinct sub-distribution. Since expert demonstrations typically represent an aggregated distribution rather than explicitly capturing individual variations, the actual target distribution can be seen as a mixture of preference distributions,
\begin{equation}
    P_{\mathrm{data}}(x) = \sum_{i}w_iP_{\mathrm{pref}}^i(x)
\end{equation}
where $P_{\mathrm{pref}}^i(x)$ represents the trajectory distribution of the $i$-th driving style, and $w_i$ is its mixture weight. Training solely on $P_{\mathrm{data}}(x)$ may cause the generative model to miss the underlying multi-modal structure, leading to suboptimal performance on trajectories outside the dominant mode of the dataset.
To align the generated trajectory distribution $P_{\theta}(x)$ with diverse human preferences, we introduce a human feedback finetuning approach that incorporates preference annotations into the training process. Specifically, given a set of human-ranked trajectories $HF=\{x_1, x_2, \dots, x_N\}$ annotated by human, we define a preference-based reward function $R(x)$ such that, 
\begin{equation}
    P_{HF}=\arg\max _{P}\mathbb{E} _{P}\left[ R \right],
\end{equation}
where $P_{HF}(x)$ represents the refined trajectory distribution incorporating human feedback. This enables the model to minimize the divergence
$D_{KL}(P_{HF}\|P_\theta)$, thereby reducing the discrepancy between the learned trajectory distribution and the human-preferred trajectory distribution.

\begin{figure}[!t]
    \centering
    \includegraphics[width=\textwidth]{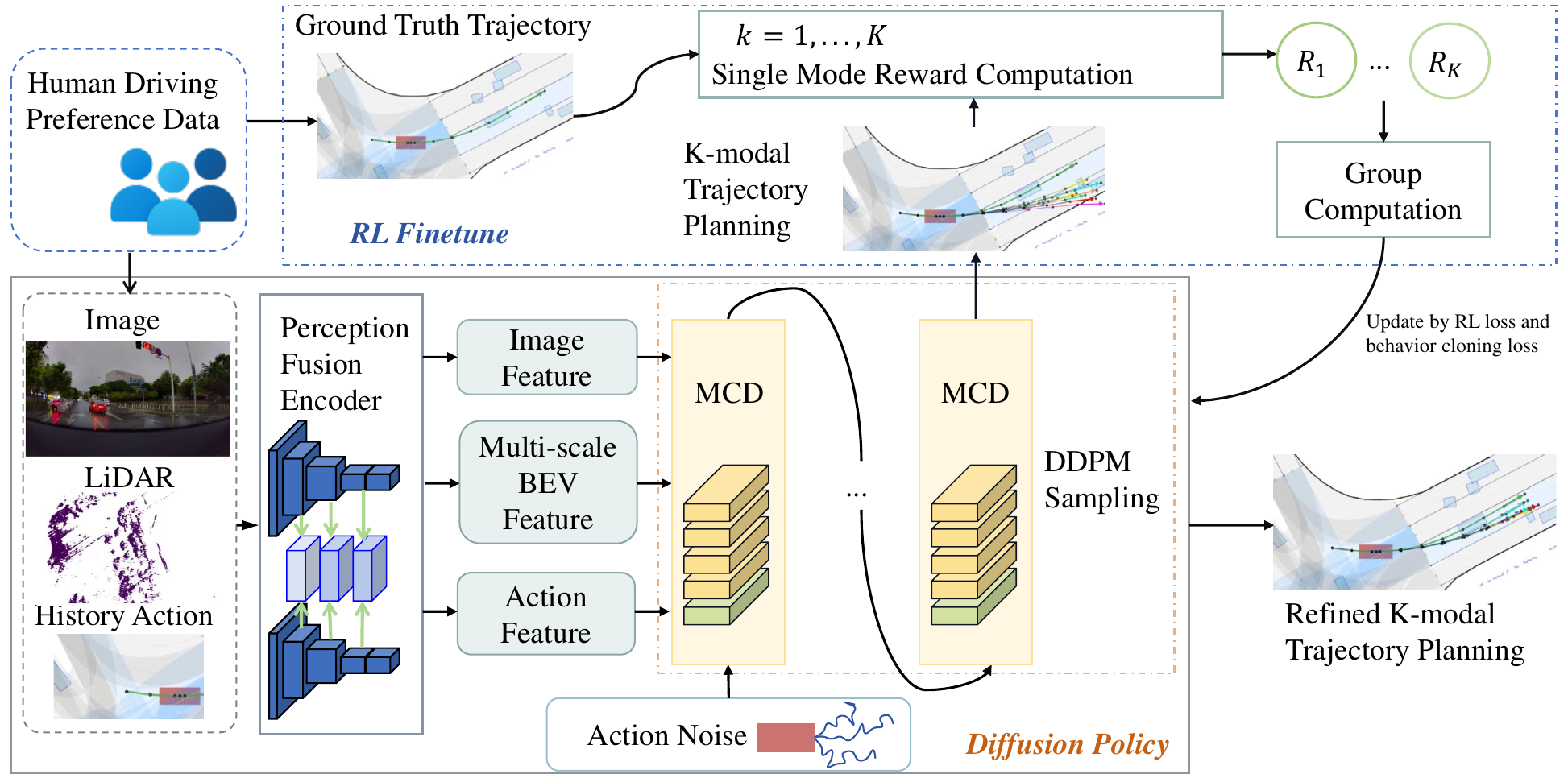}
    \caption{\textbf{Overview of \systemname\ system.} After pretraining, our diffusion policy generates $K$-modal trajectories conditioned on multi-modal inputs from human driving preference data via the Multi-Conditional Denoiser (MCD). Each trajectory mode is compared with the stylized ground truth trajectory to compute a reward. The $K$ rewards by modes are fed into group computation and used to finetune the diffusion policy through a combination of RL loss and behavior cloning loss. The system refines the multi-modal autonomous driving planning to align with the driving style knowledge in the preference data.}
    \label{fig:fig2}
    \vspace{-1em}
\end{figure}

\section{Method}

The overall architecture of our system is illustrated in Fig.~\ref{fig:fig2}. It includes two key components: diffusion policy for multi-modal trajectory generation and reinforcement learning (RL) finetuning which aligns the model with the human preference data.

\subsection{Generative Trajectory Model}
\label{sec:denoiser}
\textbf{Diffusion Model Preliminaries}
Diffusion models learn to represent the trajectory distribution $P_{\theta}(x)$ by minimizing the $KL$-divergence defined in Eq.~\ref{eq:KL}. We adopt Denoising Diffusion Probabilistic Models (DDPM) ~\citep{ho2020denoising} as the foundational framework for the denoising process. DDPM establishes a reverse denoising process that transitions from Gaussian noise $x_T$ to the noise-free state $x_0$, governed by the following transition equation: 
\begin{equation}
    x_{t-1} = \frac{1}{\sqrt{\alpha_t}}\left(x_t - \frac{1-\alpha_t}{\sqrt{1-\bar{\alpha}_t}}\epsilon_\theta(x_t,t)\right)+\sigma_tz
\end{equation}
where $z\sim\mathcal{N}(\mathbf{0},\mathbf{I})$. $t\in\{1,\dots,T\}$ denotes the noise level, and $\epsilon_\theta$ represents the noise prediction network with learnable parameters $\theta$. The Gaussian noise added to the data $x$ is according to a variance schedule $\beta_t$. Let $\alpha_t = 1 - \beta_t$, $\bar{\alpha}_t = \prod_{i=1}^t\alpha_i$ and $\sigma_t^2 = \beta_t$. The denoising model takes the current noise level and perceptual conditions as input to estimate the noise. We formalize the conditional distribution to be predicted as $P_{\theta}(x|o)$, where $o$ is the observation conditions input, including front-view images, LiDAR sensors, historical actions and ego states. The optimization objective is formulated as a gradient descent step on: 
\begin{equation}\label{slloss}
    \nabla_\theta\|\epsilon - \epsilon_\theta(\sqrt{\bar{\alpha}_t}x_0+\sqrt{1-\bar{\alpha}_t}\epsilon,t,o)\|^2
\end{equation}
where $\epsilon$ is randomly sampled from $\mathcal{N}(\mathbf{0},\mathbf{I})$.

\begin{figure}[!t]
    \centering
    \includegraphics[width=\textwidth]{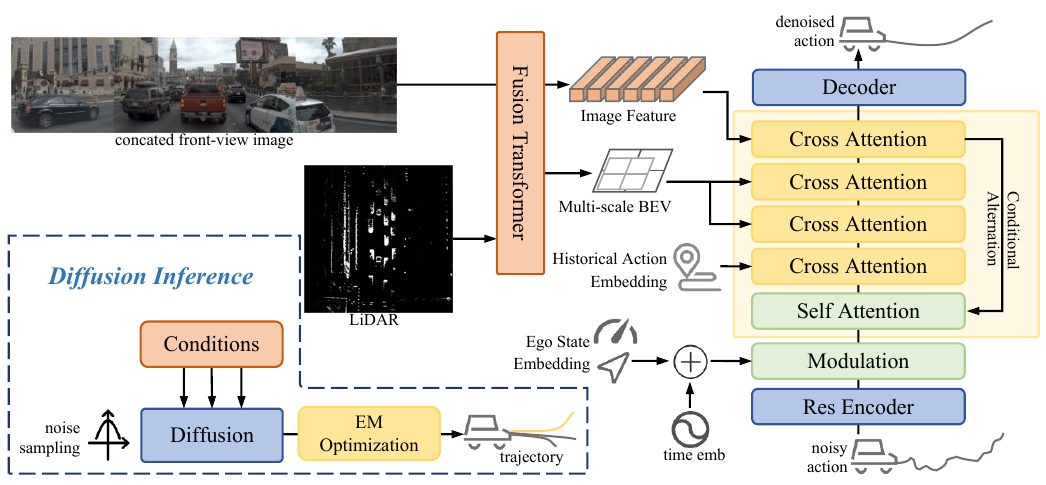}
    \caption{\textbf{The detailed architecture of multi-conditional denoiser.} The multi-modal perceptual input (image, LiDAR and actions) is encoded to different features which serve as the conditions of our transformer denoiser. During inference, multiple uniform sampling leads to multi-modal trajectory planning, which is then optimized to the final result by an EM algorithm.}
    \label{fig3}
    \vspace{-1em}
\end{figure}

\textbf{Multi-Conditional Denoiser}
To solve the complex autonomous driving problem, we design a denoising transformer to generate trajectories conditioned on multi-modal perceptions, including front-view images, LiDAR data and historical states, as shown in Fig.~\ref{fig3}. Before introducing Gaussian noise into the model input, the ego-centric trajectory is initially projected into the action space to mitigate heteroscedasticity along the trajectory timeline. The projection is expressed as $\hat{x}_l =s_l-s_{l-1}$ and $\hat{x}_0 = \boldsymbol{0}$, where $\hat{x}_l$ represents the agent's action at timestep $l$, and $l$ ranges from $1$ to $L$. This mapping ensures a reversible relationship between the two spaces, allowing the trajectory to be readily derived by accumulating actions. The subsequent noisy action is encoded by a MLP with residual, followed by modulation with state and time embeddings. Our model integrates multiple contextual conditions to guide the denoising process. Inspired by Transfuser ~\citep{chitta2022transfuser}, the concatenated front-view image and LiDAR are primarily encoded by their backbones, respectively, interleaved with fusion transformers to provide visual and BEV conditions. The intermediate fused feature maps are further upscaled to serve as a higher-resolution BEV condition. These conditions interact with the noisy action through cross attention blocks over multiple iterations, ultimately producing the predicted noise, which is then utilized in the diffusion process. During inference, the model generates a pool of candidates. An EM-algorithm based post-processing methods inspired by \citet{DBLP:journals/corr/abs-2111-14973} is used to obtain the final planning trajectory out of inference samples, detailed in Appendix~\ref{app:EM}. Notably, this approach is entirely \textbf{free} from the use of anchors or vocabulary-driven mechanisms.

\subsection{Driving Preferences Alignment}
\label{sec4.3}

To acquire quantified human driving preference, we construct dataset and train a reward model $r_{\phi}$ to evaluate the generated trajectory styles. Based on this model, we develop the Group Reward Policy Optimization for Diffusion Perference finetuning (DPGRPO) algorithm, ensuring that the generated trajectories are aligned with human preference while regressing toward practical form.

\textbf{Training Reward Model with Semi-synthetic Data} We construct a human driving preference dataset $\mathcal{D}_p=\{(o^{(j)};\mathbf{x}_c^{(j)})\}^n_{j=1}$ by mining critical scenarios from takeover data and identifying distinct driving styles. The $o$ represents an observation in a driving scenario and the $\mathbf{x}_c$ represents the chosen trajectory aligned with human preference. Based on this dataset, we develop a semi-synthetic dataset $\mathcal{D}_R=\{(o^{(j)};\mathbf{x}_c^{(j)},\mathbf{x}_i^{(j)})\}^{m}_{j=1}$ for reward model training, where the ignored trajectory $\mathbf{x}_i$ is synthesized from our pretrained model and represents normal trajectory. The details of preference data collection is presented in Appendix~\ref{app:datacollect} and the details of semi-synthetic dataset construction are presented in Appendix~\ref{app:ae}.

We employ the same denoising transformer structure described in Section~\ref{sec:denoiser} as the encoder of the reward model $r_{\phi}$, processing generated trajectory $\mathbf{x}$ and observation information $o$ including camera feature, lidar feature and past trajectory. For the decoder, we directly convert this encoded representation into scalar rewards using a multilayer perceptron.

We implement a Bradley-Terry ranking loss \citep{bradley1952rank} with margin as our reward model loss function, defined as:

\begin{equation}
\mathcal{L}^\phi
\;=\;
\mathbb{E}_{(o;\mathbf{x}_c,\mathbf{x}_i)\sim\mathcal{D}_R}
\!\Bigl[
 -\log \sigma\!\bigl(r_\phi(o,\mathbf{x}_c) -r_\phi(o,\mathbf{x}_i)\bigr)
 \;+\;
 \bigl[m - (r_\phi(o,\mathbf{x}_c) -r_\phi(o,\mathbf{x}_i)\bigr]_{+}
\Bigr],
\label{eq:pairwise-expected-loss}
\end{equation}
where $\sigma$ represents the sigmoid function, $\left[\cdot\right]_{+}$ is equivalent to $\max(0,\cdot)$ and $m$ is margin constant.

\begin{algorithm}[!t]\small
\caption{DPGRPO}\label{alg:DPGRPO}

\textbf{Input:} diffusion policy $\pi_\theta$, 
    dataset of the target driving style $\mathcal{D}_p$,
    group size $K$. \\
\textbf{Output:} finetuned policy $\pi_\theta$.
\begin{algorithmic}[1]
    \State Initialize reference policy $\pi_\theta^{\mathrm{ref}}\!\leftarrow\!\pi_\theta$
    \While{\textbf{not} converged}
        \State Sample $(o,z)\sim p(o)\times\mathcal N(\mathbf0,\mathbf I)$
        \State Sample a trajectory group $\mathcal G=\{\tau^{(k)}\}_{k=1}^{K}\sim\mathcal P_\theta(o,z)$
        \For{$k=1,\dots,K$}
            \State Compute return $r_k \leftarrow r_\phi\!\bigl(o,\mathbf x_0^{(k)}\bigr)$
        \EndFor
        \State Compute estimated group relative advantages $\hat A_{\mathrm{gr}}^{k}$ via Eq.~\ref{gradv}
        \State Compute RL loss $\mathcal L_{\mathrm{RL}}^\theta$ via Eq.~\ref{rlloss}
        \State Sample a reference trajectory $\tau^{\mathrm{ref}}\sim\mathcal P_{\mathrm{ref}}$
        \State Compute BC loss $\mathcal L_{\mathrm{BC}}^\theta$ via Eq.~\ref{bcloss}
        \State Update $\pi_\theta$ by minimizing the combined loss $\mathcal L^\theta$ of Eq.~\ref{totalloss}
    \EndWhile
    \State Apply short supervised refresh with $\mathcal{D}_p$
    \State \Return converged policy $\pi_\theta$
\end{algorithmic}
\end{algorithm}

\textbf{Finetuning Trajectory Model with Preference RL}
We formulate the diffusion denoising process as a finite-horizon Markov Decision Process (MDP), following prior work ~\citep{DDPO, DPPO}, where the diffusion planner serves as a stochastic policy that generates trajectories $\mathbf{x}$ conditioned on observations $o$ ; and the reward $r$ is directly given by $r=r_\phi(o,\mathbf{x})$.

To streamline the finetuning process, we adopt and modify GRPO method introduced by ~\citet{deepseek} and calculate the estimated group relative advantage with group size $K$ as 
\begin{equation}\label{gradv}
\hat{A}_{\mathrm{gr}}^{(k)}=\frac{r_k-\mathrm{mean}\left( \left\{ r_1,r_2,\cdots ,r_K \right\} \right)}{\mathrm{std}\left( \left\{ r_1,r_2,\cdots ,r_K \right\} \right)}
\end{equation}

Since the reward signal is provided exclusively at the terminal trajectory, the RL loss is formulated as

\begin{equation}\label{rlloss}
\begin{aligned}
\mathcal{L}_{RL}^\theta=&-\mathbb{E} _{\mathcal{G}\sim\mathcal{P}_\theta}{\left[\frac{1}{KT}\sum^K_{k=1}\sum^T_{t=1}{\log \pi _{\theta}\left( a_{t}^{(k)}\mid \psi_{t}^{(k)} \right)\gamma^{T-1-t} \hat{A}_{\mathrm{gr}}^{(k)}} \right]}
\end{aligned},
\end{equation}


where $\mathcal{G}=\{\tau^{(1)},\dots,\tau^{(K)}| \tau^{(k)}=(\psi_0^{(k)},a_0^{(k)},\dots,\psi_{T-1}^{(k)},a_{T-1}^{(k)},\psi_{T}^{(k)})\}$ is a denoising trajectory group with the same observation $o$ and noise $z$. Here $\mathcal{P}_\theta$ is a mixed sampling distribution and $\gamma\in[0,1]$ is a discount factor.

To curb overfitting on the small RL train set, we augment the RL objective with a behavior-cloning regularizer that keeps the updated policy close to a frozen pretrained reference. Specifically, following the approach outlined in ~\citet{DPPO}, the BC loss function is designed as
\begin{equation}\label{bcloss}
\mathcal{L} _{BC}^\theta=-\mathbb{E} _{\tau\sim\mathcal{P}_{ref}}\left[\frac{1}{T}\sum_{t=0}^{T-1}{\log \pi _{\theta}\left( a_t\mid \psi _t \right)}\,\, \right],
\end{equation}
where we sample state-action sequences $\tau$ using freezed initial model $\pi_\mathrm{ref}$ to contruct mixed sampling distribution $\mathcal{P}_{ref}$ and encourage the finetuned model $\pi_\theta$ to maintain a high probability of producing the same actions as $\pi_\mathrm{ref}$. The combined loss with weight coefficient $\alpha\geq 0$ is 
\begin{equation}\label{totalloss}
\mathcal{L}^\theta=\mathcal{L} _{RL}^\theta+\alpha\mathcal{L} _{BC}^\theta.
\end{equation}
After RL converges, we adopt a short supervised refresh that applies a few additional supervised learning epochs on dataset $\mathcal{D}_p$ to pull the policy back and erase residual drift. We employ the same loss function of Eq.~\ref{slloss}. The DPGRPO algorithm is summarized in Algorithm~\ref{alg:DPGRPO}, and more RL modeling details can be found in Appendix~\ref{app:rl}.

\section{Experiments}
\subsection{Experimental Setup}
\label{sec:expset}
\textbf{Dataset and Metrics} We conduct experiments on three datasets: NavSim ~\citep{dauner2024navsim}, Internal Normal dataset for pretraining, and Internal Preference dataset for finetuning. On the NavSim benchmark, our generative trajectory model is evaluated following the official Predictive Driver Model Score (PDMS)~\citep{dauner2024navsim}, which quantifies driving capacity by aggregating subscores for multiple objectives such as progress and comfort. On Internal Normal dataset, we pretrain diffusion policy with three distinct scales of training data, sampling frames at 2Hz to ensure temporal coverage. Internal Preference dataset, constructed in Appendix~\ref{app:datacollect}, has two splits, which capture ``aggressive'' and ``defensive'' driving behaviors through human annotations, respectively. On Internal dataset, model performance is assessed using metrics including minADE, meanADE, minFDE and meanFDE.


\textbf{Human Preference Evaluation} We propose human preference judgments to evaluate trajectory output styles. Evaluators compare trajectory pairs under identical contexts and select the one that better aligns with the intended style. Based on this, we develop a systematic human evaluation framework and introduce the \textbf{Better or Equal Rate (BOE)}—the proportion of instances where one model is rated better than or equal to another (see Appendix~\ref{app:boe}). This metric more precisely captures the evaluation objective centered on human preferences, which are inherently difficult to quantify.

The capacities of Internal dataset are shown in Appendix~\ref{app:data} and more implementation details of our generative model, reward model and RL finetuning are presented in Appendix~\ref{app:imp}.

\subsection{Main Results}

\subsubsection{NavSim Benchmark}

\begin{table}[!htbp]
    \renewcommand\arraystretch{1.2}
    \centering
    \caption{\textbf{Performance on \textit{navtest} with closed-loop metrics.} ``AR'' represents auto-regressive training paradigm. ``A/V'' indicates the necessity of using anchors or vocabulary. Results using PDMS selector show the upper bound of TrajHF, marked in gray. All metrics are sourced from the cited paper. }
    \resizebox{\linewidth}{!}{
    \begin{tabular}{lccccccc>{\columncolor{gray!20}}c}
        \hline
        \hline
        \textbf{Method}  &Paradigm &$A/V$ &$NC\uparrow$ & $DAC\uparrow$ & $EP\uparrow$ & $TTC\uparrow$  & $C\uparrow$ &$PDMS\uparrow$ \\     
        \hline
        Transfuser ~\citep{chitta2022transfuser}&AR &$\times$     &97.8 & 92.6 & 78.9 & 92.9  & 100 & 83.9 \\
        Hydra-MDP ~\citep{li2024hydra} &AR &$\checkmark$ & 98.3 & 96.0 & 78.7 & 94.6 & 100 & 86.5 \\
        DiffusionDrive ~\citep{diversity}  &Diffusion &$\checkmark$ &98.2 & 96.2 & 82.2 & 94.7 & 100 & 88.1 \\
        GoalFlow ~\citep{xing2025goalflow}&Diffusion &$\checkmark$ &98.4 & 98.3 & 85.0 & 94.6  & 100 & 90.3 \\
        \hline
        TrajHF (Single sample)  &Diffusion &$\times$     &96.3 & 96.0 & 83.1 & 91.5  & 100 & 86.4 \\
        \textbf{TrajHF (EM)}  &Diffusion &$\times$     &96.6 & 96.6 & 84.5 & 92.1  & 100 & 87.6\\
        \textcolor{gray}{TrajHF (PDMS selector)}  &\textcolor{gray}{Diffusion} &\textcolor{gray}{$\times$}     &\textcolor{gray}{99.2} & \textcolor{gray}{99.1} & \textcolor{gray}{90.2} & \textcolor{gray}{97.7}  & \textcolor{gray}{100} & \textcolor{gray}{94.3} \\
        \hline
        \hline \\
    \end{tabular}}
    \label{table:leaderboard}
    \vspace{-1em}
\end{table}

Table~\ref{table:leaderboard} reports the results of our model compared to other state-of-the-art models on the NavSim benchmark. We pretrain and finetune TrajHF on NavSim trainval split and utilize EM optimization from 15 candidate samples, detailed in Appendix~\ref{app:imp}. Our approach demonstrates good performance across multiple metrics, including the comprehensive PDMS metric. Moreover, the architecture we propose does not rely on predefined anchors or vocabularies. This design covers a flexible and plausible trajectory planning distribution demonstrated by the results utilizing PDMS selector, which supports the downstream RLHF module. For fairness, we do not use Internal dataset for pretraining, and all methods adopt both camera and LiDAR as input.

\subsubsection{Human Preference Finetuning}\label{sec:rlft}
\begin{table}[!htbp]
    \renewcommand\arraystretch{1.2}
    \centering
    \caption{\textbf{Performance of finetuning on Internal test splits with open-loop metrics.} ``Diversity'' is calculated as defined in ~\citet{diversity}. ``BOE'' is calculated over ``w/o FT'' model. ``Aggr.'' and ``Defen.'' refer to the aggressive and defensive datasets used for applying the respective methods. For ``Normal'' test split, baseline model w/o FT is not considered in comparison, marked in gray.}
    \resizebox{\linewidth}{!}{
    \begin{tabular}{lccccccc}
        \hline
        \textbf{Method} &Test Split & minADE$\downarrow$ & meanADE$\downarrow$ & minFDE$\downarrow$ & meanFDE$\downarrow$ & Diversity$\uparrow$ & BOE $\uparrow$  \\  
        \hline
        w/o FT &Aggressive &0.7546         &1.5534         &2.0054          &4.3515         &0.4419          &- \\
        Aggr. SL     &Aggressive &0.5416 &\textbf{1.4485}          &1.3147 &\textbf{3.9245 }         &0.5743 &0.6577 \\
        Aggr. RL     &Aggressive &\textbf{0.5304}          &1.5036 &\textbf{1.2788 }         &4.1039 &\textbf{0.6051 }   &\textbf{0.7660} \\ 
        \hline
        w/o FT &Defensive &0.5538         &\textbf{1.4419}          &1.5854          &\textbf{4.3763}          &0.3741          &- \\
        Defen. SL     &Defensive &0.3175 &1.5317         &0.8351 &4.5239        &0.3714 &0.7430 \\
        Defen. RL     &Defensive &\textbf{0.3101}          &1.5309 &\textbf{0.8092 }         &4.5603 &\textbf{0.3900}    &\textbf{0.7922} \\ 
        \hline
        w/o FT &Normal &\textcolor{gray}{0.2330} &\textcolor{gray}{0.7734}         &\textcolor{gray}{0.5454}   &\textcolor{gray}{2.1207}          &\textcolor{gray}{0.3284}         &- \\
        Aggr. SL     &Normal &0.4612          &1.3290          &1.2850            &3.8658          &0.5276  &- \\
        Aggr. RL     &Normal &\textbf{0.3777}          &\textbf{1.3034} &\textbf{1.0247}           &\textbf{3.8057} &\textbf{0.5575}    &-\\
        \hline
        w/o FT &Normal &\textcolor{gray}{0.2330} &\textcolor{gray}{0.7734}         &\textcolor{gray}{0.5454} &\textcolor{gray}{2.1207}          &\textcolor{gray}{0.3284}         &- \\
        Defen. SL     &Normal &0.3675          &3.0225          &0.9428           &8.7625          &\textbf{0.3811}  &- \\
        Defen. RL     &Normal &\textbf{0.3361}          &\textbf{2.5843} &\textbf{0.8537}          &\textbf{7.4631} &0.3773    &-\\
        \hline
    \end{tabular}}
    \vspace{0.6em}
    \label{table:ft}
    \vspace{-1.5em}
\end{table}

We evaluate RL finetuning method and supervised learning (SL) method on Internal normal and preference test splits. SL employs the same loss function of Eq.~\ref{slloss} as in the pretraining phase, while RL utilizes Algorithm~\ref{alg:DPGRPO}. The comparative experimental results are summarized in Table~\ref{table:ft}. 

Both SL and RL finetuning effectively adapt driving styles toward greater aggressiveness or defensiveness while reducing distance errors. When comparing methods, RL consistently outperforms SL with higher \textbf{BOE} score and lower minADE and minFDE values, demonstrating superior best-case performance. \textbf{BOE} especially shows the effectiveness of \systemname\ aligning with distinct personalized driving style. Additionally, RL generates trajectories with higher diversity metrics across preference datasets, producing more varied yet feasible driving behaviors.
Furthermore, RL demonstrates greater practical utility by maintaining excellent performance on normal test splits despite preference-based finetuning, surpassing SL in nearly all metrics and preference indicators. This highlights RL's robust generalization capabilities across varied driving scenarios without significant performance degradation.
Although SL achieves marginally better mean error metrics in some cases, this advantage likely stems from the inherent variance in RL method.
\begin{figure}[!t]
    \centering
    \includegraphics[width=\textwidth]{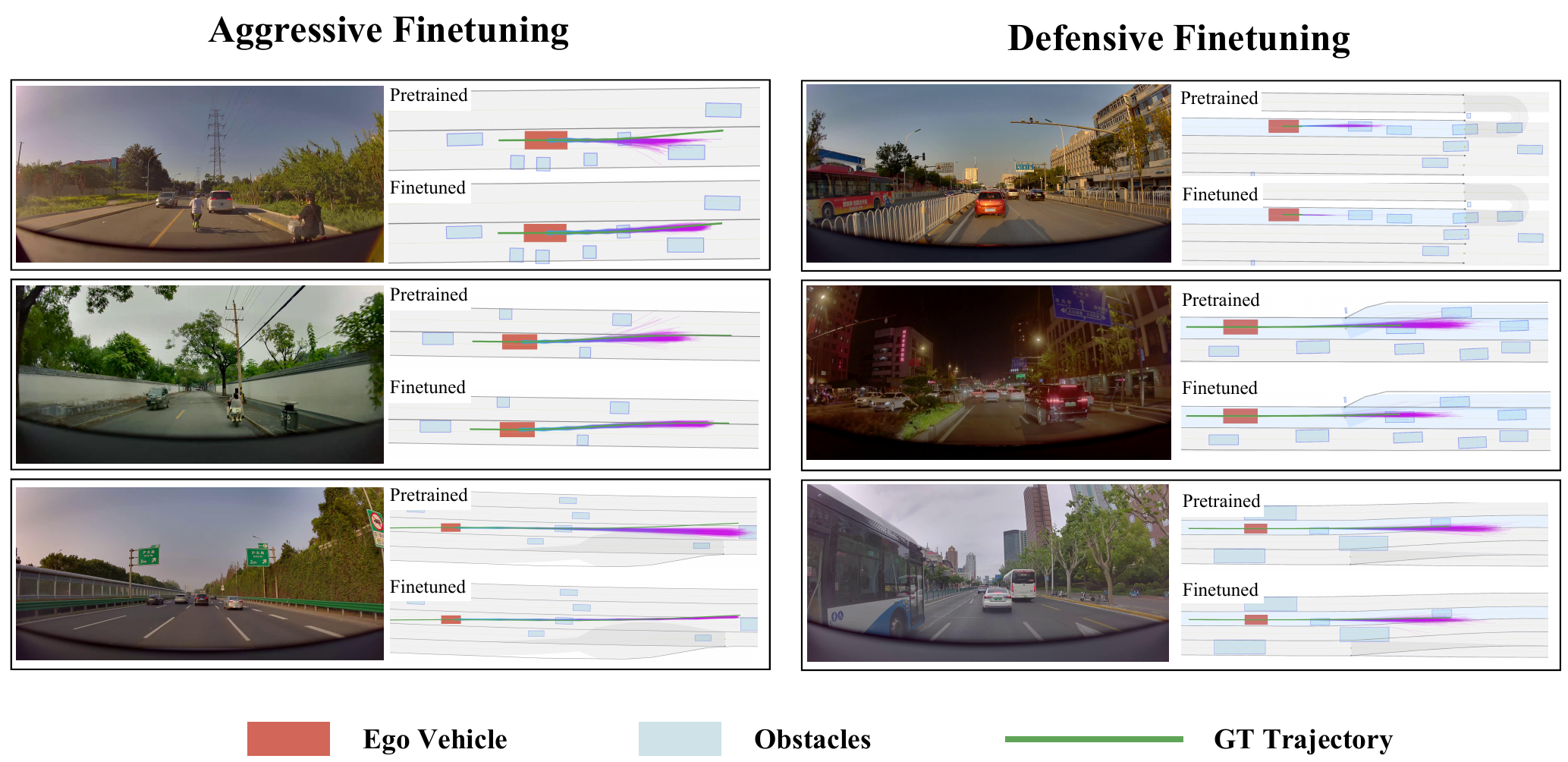}
    \caption{\textbf{Qualitative comparison of pretrained vs. 
 RL finetuned diffusion policies.} On the right side of each case, the upper half and the lower half BEV images show the planned trajectories of the pretrained policy and the RL finetuned policy, respectively.}
    \label{fig:compare}
    \vspace{-1em}
\end{figure}

We conduct qualitative analysis to assess the effect of preference-aligned finetuning. As shown in Fig.~\ref{fig:compare}, the finetuned model demonstrates improved context awareness and adaptability—proactively overtaking in complex scenarios for efficiency, or decelerating in ambiguous or risky situations for safety. These behaviors are absent in the pretrained policy, indicating that RL method effectively guides the model toward more realistic and human-aligned driving decisions. More detailed description and analysis for each visualized case are provided in Appendix~\ref{app:qual}.

Fig.~\ref{fig:velo} presents the velocity distributions for three driving styles from both model predictions and ground truth of test splits. The results show distinct velocity profiles across styles, demonstrating a positive correlation between stylization and vehicle speed. Specifically, the aggressive style shows higher probability density in high-speed regimes (24-40 m/s) than other styles. The finetuned model's predictions align well with the ground truth velocity distributions, demonstrating the effectiveness of our \systemname\ framework to align the trajectory planning with personalized driving styles.

\begin{figure}[!t]
    \centering
    
    \begin{minipage}[c]{0.49\textwidth}
        \centering
        \includegraphics[width=\linewidth]{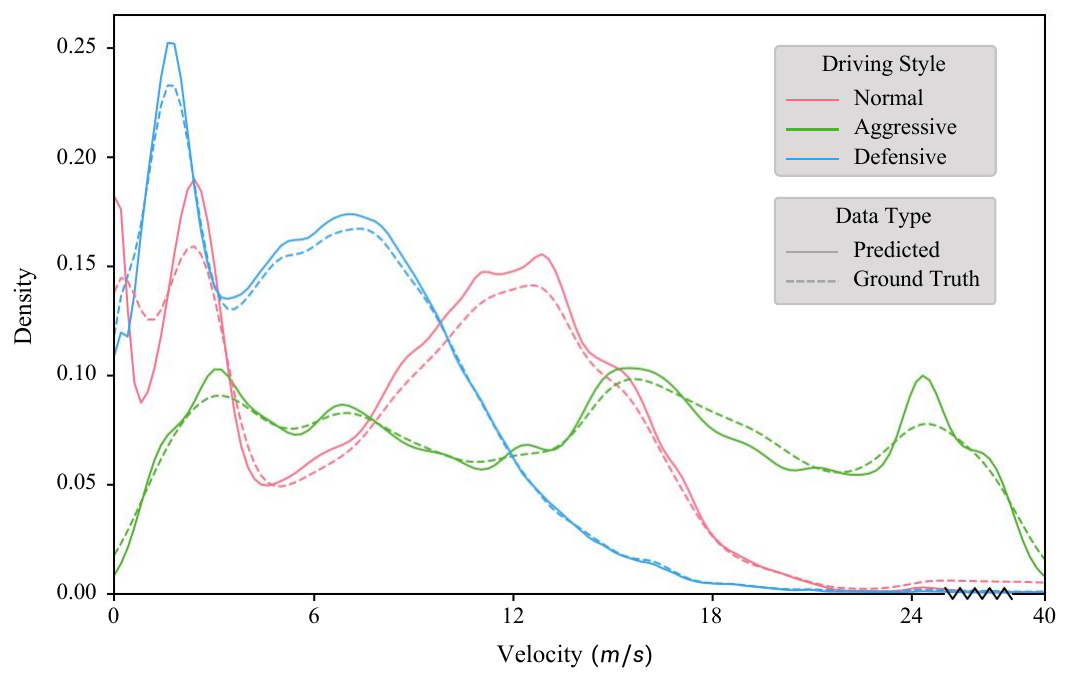}
        \caption{\textbf{Velocity distribution across driving styles.} Comparison of speed density profiles between model predictions (solid lines) and ground truth (dashed lines) for different styles.}
        \label{fig:velo}
    \end{minipage}
    \hfill
    \begin{minipage}[c]{0.49\textwidth}
        \footnotesize
        \centering
        \captionof{table}{\textbf{Impact of Scaling the Training Set Size.} Here miA, meA, miF, meF and Div indicate minADE, meanADE, minFDE, meanFDE and Diversity, respectively.}
        \begin{tabular}{lccccc}
            \hline
            \hline
            \textbf{Scale}  & miA$\downarrow$ & meA$\downarrow$ & miF$\downarrow$ & meF$\downarrow$ & Div$\uparrow$ \\     
            \hline
            810K  & 0.40          & 1.09         & 0.87          & 2.67         & \textbf{0.28} \\
            3M & 0.39          & 1.08        & 0.84          & 2.70         & 0.26 \\
            6M & \textbf{0.37} &\textbf{1.04} & \textbf{0.81} &\textbf{2.66} &0.26 \\
            \hline
            \hline \\
        \end{tabular} 
        \label{table:datascale}
  \end{minipage}
  \vspace{-1em}
\end{figure}

\subsection{Ablation Study}\label{sec:ablation}

Table~\ref{table:datascale} illustrates the impact of training data scale in pretraining. The results are evaluated on the Internal normal test split. We observe that as the dataset size grows, the model exhibits non-marginal improvements in distance error metrics. However, this comes at the cost of reduced diversity, indicating that the diffusion policy learns a more concentrated distribution with larger data volumes.

In Appendix~\ref{app:sup}, we report the ablation studies for Algorithm~\ref{alg:DPGRPO}. These analyses include: (i) PPO-based variants derived from prior work and widely used approach (Appendix~\ref{app:ppov}); (ii) the effect of varying behavior cloning (BC) loss weights (Appendix~\ref{app:bc}); and (iii) the influence of data scaling strategies (Appendix~\ref{app:datascale}). Collectively, these experiments substantiate the effectiveness and robustness of our proposed DPGRPO fine-tuning approach.

\section{Conclusion}
\label{sec:con}
In this work, we introduce \systemname, a human feedback-driven finetuning framework for generative trajectory models that aligns autonomous driving behaviors with diverse personal driving styles. By combining a DDPM-based multi-modal planner with reinforcement learning, \systemname\ enables safe, feasible, and personalized trajectory generation. Experimental results demonstrate that \systemname\ preserves key motion planning constraints while aligning to the specific human driving style. By leveraging human feedback as a supervisory signal, our approach addresses the limitations of traditional imitation learning, offering a more flexible and human-aligned trajectory generation paradigm. 

\textbf{Limitations}
The framework depends on human annotations, which can be subjective, inconsistent, and costly. Its generalization capacity is tied to the diversity and quality of collected preferences. Future work will explore auto-labeling via takeover data and scaling to broader contexts with richer social cues.

\textbf{Broader Impact}
By generating more natural and socially aligned behaviors, our method can improve user trust and acceptability in autonomous vehicles. The framework may also extend to other domains requiring preference-aligned decisions. However, personalization in safety-critical systems introduces risks such as reinforcing unsafe or biased behaviors, requiring careful deployment and oversight.

\newpage
\bibliographystyle{unsrtnat}
\bibliography{neurips_2025}

\begin{thebibliography}{42}
\providecommand{\natexlab}[1]{#1}
\providecommand{\url}[1]{\texttt{#1}}
\expandafter\ifx\csname urlstyle\endcsname\relax
  \providecommand{\doi}[1]{doi: #1}\else
  \providecommand{\doi}{doi: \begingroup \urlstyle{rm}\Url}\fi

\bibitem[Zhao et~al.(2021)Zhao, Gao, Lan, Sun, Sapp, Varadarajan, Shen, Shen, Chai, Schmid, et~al.]{zhao2021tnt}
Hang Zhao, Jiyang Gao, Tian Lan, Chen Sun, Ben Sapp, Balakrishnan Varadarajan, Yue Shen, Yi~Shen, Yuning Chai, Cordelia Schmid, et~al.
\newblock Tnt: Target-driven trajectory prediction.
\newblock In \emph{Conference on Robot Learning}, pages 895--904. PMLR, 2021.

\bibitem[Gu et~al.(2021)Gu, Sun, and Zhao]{gu2021densetnt}
Junru Gu, Chen Sun, and Hang Zhao.
\newblock Densetnt: End-to-end trajectory prediction from dense goal sets.
\newblock In \emph{Proceedings of the IEEE/CVF International Conference on Computer Vision}, pages 15303--15312, 2021.

\bibitem[Xing et~al.(2025)Xing, Zhang, Hu, Jiang, He, Zhang, Long, and Yin]{xing2025goalflow}
Zebin Xing, Xingyu Zhang, Yang Hu, Bo~Jiang, Tong He, Qian Zhang, Xiaoxiao Long, and Wei Yin.
\newblock Goalflow: Goal-driven flow matching for multimodal trajectories generation in end-to-end autonomous driving.
\newblock \emph{arXiv preprint arXiv:2503.05689}, 2025.

\bibitem[Jiang et~al.(2023{\natexlab{a}})Jiang, Cornman, Park, Sapp, Zhou, Anguelov, et~al.]{jiang2023motiondiffuser}
Chiyu Jiang, Andre Cornman, Cheolho Park, Benjamin Sapp, Yin Zhou, Dragomir Anguelov, et~al.
\newblock Motiondiffuser: Controllable multi-agent motion prediction using diffusion.
\newblock In \emph{Proceedings of the IEEE/CVF conference on computer vision and pattern recognition}, pages 9644--9653, 2023{\natexlab{a}}.

\bibitem[Jiang et~al.(2024)Jiang, Bai, Cornman, Davis, Huang, Jeon, Kulshrestha, Lambert, Li, Zhou, et~al.]{jiang2024scenediffuser}
Max Jiang, Yijing Bai, Andre Cornman, Christopher Davis, Xiukun Huang, Hong Jeon, Sakshum Kulshrestha, John Lambert, Shuangyu Li, Xuanyu Zhou, et~al.
\newblock Scenediffuser: Efficient and controllable driving simulation initialization and rollout.
\newblock \emph{Advances in Neural Information Processing Systems}, 37:\penalty0 55729--55760, 2024.

\bibitem[Ngiam et~al.(2021)Ngiam, Caine, Vasudevan, Zhang, Chiang, Ling, Roelofs, Bewley, Liu, Venugopal, et~al.]{ngiam2021scene}
Jiquan Ngiam, Benjamin Caine, Vijay Vasudevan, Zhengdong Zhang, Hao-Tien~Lewis Chiang, Jeffrey Ling, Rebecca Roelofs, Alex Bewley, Chenxi Liu, Ashish Venugopal, et~al.
\newblock Scene transformer: A unified architecture for predicting multiple agent trajectories.
\newblock \emph{arXiv preprint arXiv:2106.08417}, 2021.

\bibitem[Sun et~al.(2023)Sun, Zhang, Ma, Shi, Li, Luo, Wang, Xu, Cao, and Zhao]{sun2023large}
Qiao Sun, Shiduo Zhang, Danjiao Ma, Jingzhe Shi, Derun Li, Simian Luo, Yu~Wang, Ningyi Xu, Guangzhi Cao, and Hang Zhao.
\newblock Large trajectory models are scalable motion predictors and planners.
\newblock \emph{arXiv preprint arXiv:2310.19620}, 2023.

\bibitem[Dauner et~al.(2024)Dauner, Hallgarten, Li, Weng, Huang, Yang, Li, Gilitschenski, Ivanovic, Pavone, et~al.]{dauner2024navsim}
Daniel Dauner, Marcel Hallgarten, Tianyu Li, Xinshuo Weng, Zhiyu Huang, Zetong Yang, Hongyang Li, Igor Gilitschenski, Boris Ivanovic, Marco Pavone, et~al.
\newblock Navsim: Data-driven non-reactive autonomous vehicle simulation and benchmarking.
\newblock \emph{Advances in Neural Information Processing Systems}, 37:\penalty0 28706--28719, 2024.

\bibitem[Codevilla et~al.(2018)Codevilla, M{\"u}ller, L{\'o}pez, Koltun, and Dosovitskiy]{codevilla2018end}
Felipe Codevilla, Matthias M{\"u}ller, Antonio L{\'o}pez, Vladlen Koltun, and Alexey Dosovitskiy.
\newblock End-to-end driving via conditional imitation learning.
\newblock In \emph{2018 IEEE international conference on robotics and automation (ICRA)}, pages 4693--4700. IEEE, 2018.

\bibitem[Hu et~al.(2023)Hu, Yang, Chen, Li, Sima, Zhu, Chai, Du, Lin, Wang, et~al.]{hu2023planning}
Yihan Hu, Jiazhi Yang, Li~Chen, Keyu Li, Chonghao Sima, Xizhou Zhu, Siqi Chai, Senyao Du, Tianwei Lin, Wenhai Wang, et~al.
\newblock Planning-oriented autonomous driving.
\newblock In \emph{Proceedings of the IEEE/CVF conference on computer vision and pattern recognition}, pages 17853--17862, 2023.

\bibitem[Gu et~al.(2023)Gu, Hu, Zhang, Chen, Wang, Wang, and Zhao]{gu2023vip3d}
Junru Gu, Chenxu Hu, Tianyuan Zhang, Xuanyao Chen, Yilun Wang, Yue Wang, and Hang Zhao.
\newblock Vip3d: End-to-end visual trajectory prediction via 3d agent queries.
\newblock In \emph{Proceedings of the IEEE/CVF Conference on Computer Vision and Pattern Recognition}, pages 5496--5506, 2023.

\bibitem[Jiang et~al.(2023{\natexlab{b}})Jiang, Chen, Xu, Liao, Chen, Zhou, Zhang, Liu, Huang, and Wang]{jiang2023vad}
Bo~Jiang, Shaoyu Chen, Qing Xu, Bencheng Liao, Jiajie Chen, Helong Zhou, Qian Zhang, Wenyu Liu, Chang Huang, and Xinggang Wang.
\newblock Vad: Vectorized scene representation for efficient autonomous driving.
\newblock In \emph{Proceedings of the IEEE/CVF International Conference on Computer Vision}, pages 8340--8350, 2023{\natexlab{b}}.

\bibitem[Chen et~al.(2024)Chen, Jiang, Gao, Liao, Xu, Zhang, Huang, Liu, and Wang]{chen2024vadv2}
Shaoyu Chen, Bo~Jiang, Hao Gao, Bencheng Liao, Qing Xu, Qian Zhang, Chang Huang, Wenyu Liu, and Xinggang Wang.
\newblock Vadv2: End-to-end vectorized autonomous driving via probabilistic planning.
\newblock \emph{arXiv preprint arXiv:2402.13243}, 2024.

\bibitem[Chitta et~al.(2022)Chitta, Prakash, Jaeger, Yu, Renz, and Geiger]{chitta2022transfuser}
Kashyap Chitta, Aditya Prakash, Bernhard Jaeger, Zehao Yu, Katrin Renz, and Andreas Geiger.
\newblock Transfuser: Imitation with transformer-based sensor fusion for autonomous driving.
\newblock \emph{IEEE transactions on pattern analysis and machine intelligence}, 45\penalty0 (11):\penalty0 12878--12895, 2022.

\bibitem[Ye et~al.(2023)Ye, Jing, Hu, Huang, Gao, Li, Wang, Guo, Xiao, Mao, et~al.]{ye2023fusionad}
Tengju Ye, Wei Jing, Chunyong Hu, Shikun Huang, Lingping Gao, Fangzhen Li, Jingke Wang, Ke~Guo, Wencong Xiao, Weibo Mao, et~al.
\newblock Fusionad: Multi-modality fusion for prediction and planning tasks of autonomous driving.
\newblock \emph{arXiv preprint arXiv:2308.01006}, 2023.

\bibitem[Li et~al.(2024)Li, Li, Wang, Lan, Yu, Ji, Li, Zhu, Kautz, Wu, et~al.]{li2024hydra}
Zhenxin Li, Kailin Li, Shihao Wang, Shiyi Lan, Zhiding Yu, Yishen Ji, Zhiqi Li, Ziyue Zhu, Jan Kautz, Zuxuan Wu, et~al.
\newblock Hydra-mdp: End-to-end multimodal planning with multi-target hydra-distillation.
\newblock \emph{arXiv preprint arXiv:2406.06978}, 2024.

\bibitem[Gupta et~al.(2018)Gupta, Johnson, Fei-Fei, Savarese, and Alahi]{gupta2018social}
Agrim Gupta, Justin Johnson, Li~Fei-Fei, Silvio Savarese, and Alexandre Alahi.
\newblock Social gan: Socially acceptable trajectories with generative adversarial networks.
\newblock In \emph{Proceedings of the IEEE conference on computer vision and pattern recognition}, pages 2255--2264, 2018.

\bibitem[Fang et~al.(2022)Fang, Zhang, Zhou, Qian, and Gan]{fang2022atten}
Fang Fang, Pengpeng Zhang, Bo~Zhou, Kun Qian, and Yahui Gan.
\newblock Atten-gan: pedestrian trajectory prediction with gan based on attention mechanism.
\newblock \emph{Cognitive Computation}, 14\penalty0 (6):\penalty0 2296--2305, 2022.

\bibitem[Xu et~al.(2022)Xu, Hayet, and Karamouzas]{xu2022socialvae}
Pei Xu, Jean-Bernard Hayet, and Ioannis Karamouzas.
\newblock Socialvae: Human trajectory prediction using timewise latents.
\newblock In \emph{European Conference on Computer Vision}, pages 511--528. Springer, 2022.

\bibitem[De~Miguel et~al.(2022)De~Miguel, Armingol, and Garc{\'\i}a]{de2022vehicles}
Miguel~{\'A}ngel De~Miguel, Jos{\'e}~Mar{\'\i}a Armingol, and Fernando Garc{\'\i}a.
\newblock Vehicles trajectory prediction using recurrent vae network.
\newblock \emph{IEEE Access}, 10:\penalty0 32742--32749, 2022.

\bibitem[Huang et~al.(2024)Huang, Zhang, Vaidya, Chen, Lv, and Fisac]{huang2024versatile}
Zhiyu Huang, Zixu Zhang, Ameya Vaidya, Yuxiao Chen, Chen Lv, and Jaime~Fern{\'a}ndez Fisac.
\newblock Versatile scene-consistent traffic scenario generation as optimization with diffusion.
\newblock \emph{arXiv preprint arXiv:2404.02524}, 3, 2024.

\bibitem[Liao et~al.(2024{\natexlab{a}})Liao, Chen, Yin, Jiang, Wang, Yan, Zhang, Li, Zhang, Zhang, et~al.]{liao2024diffusiondrive}
Bencheng Liao, Shaoyu Chen, Haoran Yin, Bo~Jiang, Cheng Wang, Sixu Yan, Xinbang Zhang, Xiangyu Li, Ying Zhang, Qian Zhang, et~al.
\newblock Diffusiondrive: Truncated diffusion model for end-to-end autonomous driving.
\newblock \emph{arXiv preprint arXiv:2411.15139}, 2024{\natexlab{a}}.

\bibitem[Christiano et~al.(2017)Christiano, Leike, Brown, Martic, Legg, and Amodei]{christiano2017deep}
Paul~F Christiano, Jan Leike, Tom Brown, Miljan Martic, Shane Legg, and Dario Amodei.
\newblock Deep reinforcement learning from human preferences.
\newblock \emph{Advances in neural information processing systems}, 30, 2017.

\bibitem[Ouyang et~al.(2022)Ouyang, Wu, Jiang, Almeida, Wainwright, Mishkin, Zhang, Agarwal, Slama, Ray, et~al.]{ouyang2022training}
Long Ouyang, Jeffrey Wu, Xu~Jiang, Diogo Almeida, Carroll Wainwright, Pamela Mishkin, Chong Zhang, Sandhini Agarwal, Katarina Slama, Alex Ray, et~al.
\newblock Training language models to follow instructions with human feedback.
\newblock \emph{Advances in neural information processing systems}, 35:\penalty0 27730--27744, 2022.

\bibitem[Yu et~al.(2024)Yu, Yao, Zhang, He, Han, Cui, Hu, Liu, Zheng, Sun, et~al.]{yu2024rlhf}
Tianyu Yu, Yuan Yao, Haoye Zhang, Taiwen He, Yifeng Han, Ganqu Cui, Jinyi Hu, Zhiyuan Liu, Hai-Tao Zheng, Maosong Sun, et~al.
\newblock Rlhf-v: Towards trustworthy mllms via behavior alignment from fine-grained correctional human feedback.
\newblock In \emph{Proceedings of the IEEE/CVF Conference on Computer Vision and Pattern Recognition}, pages 13807--13816, 2024.

\bibitem[Zhang et~al.(2024)Zhang, Zheng, Chen, Jang, Li, Han, Wang, Ding, Fox, and Yao]{zhang2024grape}
Zijian Zhang, Kaiyuan Zheng, Zhaorun Chen, Joel Jang, Yi~Li, Siwei Han, Chaoqi Wang, Mingyu Ding, Dieter Fox, and Huaxiu Yao.
\newblock Grape: Generalizing robot policy via preference alignment.
\newblock \emph{arXiv preprint arXiv:2411.19309}, 2024.

\bibitem[Shao et~al.(2024)Shao, Wang, Zhu, Xu, Song, Bi, Zhang, Zhang, Li, Wu, et~al.]{shao2024deepseekmath}
Zhihong Shao, Peiyi Wang, Qihao Zhu, Runxin Xu, Junxiao Song, Xiao Bi, Haowei Zhang, Mingchuan Zhang, YK~Li, Y~Wu, et~al.
\newblock Deepseekmath: Pushing the limits of mathematical reasoning in open language models.
\newblock \emph{arXiv preprint arXiv:2402.03300}, 2024.

\bibitem[Wallace et~al.(2024)Wallace, Dang, Rafailov, Zhou, Lou, Purushwalkam, Ermon, Xiong, Joty, and Naik]{wallace2024diffusion}
Bram Wallace, Meihua Dang, Rafael Rafailov, Linqi Zhou, Aaron Lou, Senthil Purushwalkam, Stefano Ermon, Caiming Xiong, Shafiq Joty, and Nikhil Naik.
\newblock Diffusion model alignment using direct preference optimization.
\newblock In \emph{Proceedings of the IEEE/CVF Conference on Computer Vision and Pattern Recognition}, pages 8228--8238, 2024.

\bibitem[Wang et~al.(2024)Wang, Liu, Wang, and Xiong]{wang2024reinforcement}
Yuting Wang, Lu~Liu, Maonan Wang, and Xi~Xiong.
\newblock Reinforcement learning from human feedback for lane changing of autonomous vehicles in mixed traffic.
\newblock \emph{arXiv preprint arXiv:2408.04447}, 2024.

\bibitem[Sun et~al.(2024)Sun, Salami~Pargoo, Jin, and Ortiz]{Sun_2024}
Yuan Sun, Navid Salami~Pargoo, Peter Jin, and Jorge Ortiz.
\newblock Optimizing autonomous driving for safety: A human-centric approach with llm-enhanced rlhf.
\newblock In \emph{Companion of the 2024 on ACM International Joint Conference on Pervasive and Ubiquitous Computing}, UbiComp ’24, page 76–80. ACM, October 2024.
\newblock \doi{10.1145/3675094.3677588}.
\newblock URL \url{http://dx.doi.org/10.1145/3675094.3677588}.

\bibitem[Guo et~al.(2025)Guo, Yang, Zhang, Song, Zhang, Xu, Zhu, Ma, Wang, Bi, et~al.]{deepseek}
Daya Guo, Dejian Yang, Haowei Zhang, Junxiao Song, Ruoyu Zhang, Runxin Xu, Qihao Zhu, Shirong Ma, Peiyi Wang, Xiao Bi, et~al.
\newblock Deepseek-r1: Incentivizing reasoning capability in llms via reinforcement learning.
\newblock \emph{arXiv preprint arXiv:2501.12948}, 2025.

\bibitem[Ross and Bagnell(2010)]{ross2010efficient}
St{\'e}phane Ross and Drew Bagnell.
\newblock Efficient reductions for imitation learning.
\newblock In \emph{Proceedings of the thirteenth international conference on artificial intelligence and statistics}, pages 661--668. JMLR Workshop and Conference Proceedings, 2010.

\bibitem[Ho et~al.(2020)Ho, Jain, and Abbeel]{ho2020denoising}
Jonathan Ho, Ajay Jain, and Pieter Abbeel.
\newblock Denoising diffusion probabilistic models.
\newblock \emph{Advances in neural information processing systems}, 33:\penalty0 6840--6851, 2020.

\bibitem[Varadarajan et~al.(2021)Varadarajan, Hefny, Srivastava, Refaat, Nayakanti, Cornman, Chen, Douillard, Lam, Anguelov, and Sapp]{DBLP:journals/corr/abs-2111-14973}
Balakrishnan Varadarajan, Ahmed Hefny, Avikalp Srivastava, Khaled~S. Refaat, Nigamaa Nayakanti, Andre Cornman, Kan Chen, Bertrand Douillard, Chi{-}Pang Lam, Dragomir Anguelov, and Benjamin Sapp.
\newblock Multipath++: Efficient information fusion and trajectory aggregation for behavior prediction.
\newblock \emph{CoRR}, abs/2111.14973, 2021.
\newblock URL \url{https://arxiv.org/abs/2111.14973}.

\bibitem[Bradley and Terry(1952)]{bradley1952rank}
Ralph~Allan Bradley and Milton~E. Terry.
\newblock Rank analysis of incomplete block designs: I. the method of paired comparisons.
\newblock \emph{Biometrika}, 39\penalty0 (3/4):\penalty0 324--345, 1952.
\newblock ISSN 00063444, 14643510.
\newblock URL \url{http://www.jstor.org/stable/2334029}.

\bibitem[Black et~al.(2023)Black, Janner, Du, Kostrikov, and Levine]{DDPO}
Kevin Black, Michael Janner, Yilun Du, Ilya Kostrikov, and Sergey Levine.
\newblock Training diffusion models with reinforcement learning.
\newblock \emph{arXiv preprint arXiv:2305.13301}, 2023.

\bibitem[Ren et~al.(2024)Ren, Lidard, Ankile, Simeonov, Agrawal, Majumdar, Burchfiel, Dai, and Simchowitz]{DPPO}
Allen~Z Ren, Justin Lidard, Lars~L Ankile, Anthony Simeonov, Pulkit Agrawal, Anirudha Majumdar, Benjamin Burchfiel, Hongkai Dai, and Max Simchowitz.
\newblock Diffusion policy policy optimization.
\newblock \emph{arXiv preprint arXiv:2409.00588}, 2024.

\bibitem[Liao et~al.(2024{\natexlab{b}})Liao, Chen, Yin, Jiang, Wang, Yan, Zhang, Li, Zhang, Zhang, et~al.]{diversity}
Bencheng Liao, Shaoyu Chen, Haoran Yin, Bo~Jiang, Cheng Wang, Sixu Yan, Xinbang Zhang, Xiangyu Li, Ying Zhang, Qian Zhang, et~al.
\newblock Diffusiondrive: Truncated diffusion model for end-to-end autonomous driving.
\newblock \emph{arXiv preprint arXiv:2411.15139}, 2024{\natexlab{b}}.

\bibitem[Bishop and Nasrabadi(2006)]{bishop2006pattern}
Christopher~M Bishop and Nasser~M Nasrabadi.
\newblock \emph{Pattern recognition and machine learning}, volume~4.
\newblock Springer, 2006.

\bibitem[Dosovitskiy et~al.(2020)Dosovitskiy, Beyer, Kolesnikov, Weissenborn, Zhai, Unterthiner, Dehghani, Minderer, Heigold, Gelly, et~al.]{dosovitskiy2020image}
Alexey Dosovitskiy, Lucas Beyer, Alexander Kolesnikov, Dirk Weissenborn, Xiaohua Zhai, Thomas Unterthiner, Mostafa Dehghani, Matthias Minderer, Georg Heigold, Sylvain Gelly, et~al.
\newblock An image is worth 16x16 words: Transformers for image recognition at scale.
\newblock \emph{arXiv preprint arXiv:2010.11929}, 2020.

\bibitem[He et~al.(2016)He, Zhang, Ren, and Sun]{he2016deep}
Kaiming He, Xiangyu Zhang, Shaoqing Ren, and Jian Sun.
\newblock Deep residual learning for image recognition.
\newblock In \emph{Proceedings of the IEEE conference on computer vision and pattern recognition}, pages 770--778, 2016.

\bibitem[Rafailov et~al.(2023)Rafailov, Sharma, Mitchell, Manning, Ermon, and Finn]{rafailov2023direct}
Rafael Rafailov, Archit Sharma, Eric Mitchell, Christopher~D Manning, Stefano Ermon, and Chelsea Finn.
\newblock Direct preference optimization: Your language model is secretly a reward model.
\newblock \emph{Advances in neural information processing systems}, 36:\penalty0 53728--53741, 2023.

\end{thebibliography}

\newpage
\appendix
\section{Details of Preferences Alignment Method}

\subsection{Preference Data Collection}
\label{app:datacollect}

\begin{figure}[!htbp]
    \centering
    \includegraphics[width=\textwidth]{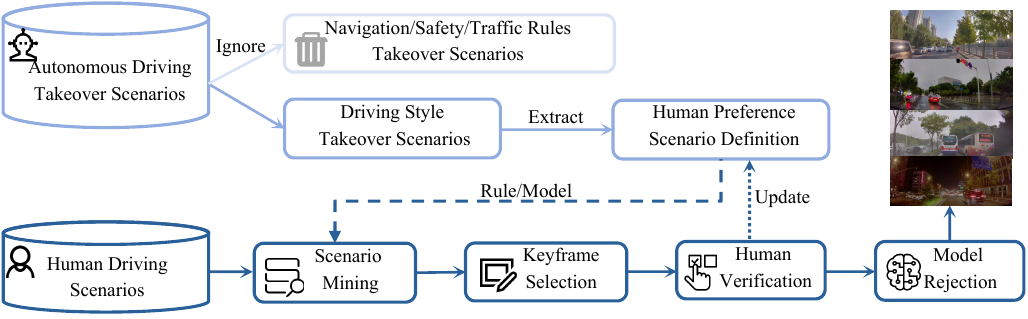}
    \caption{\textbf{The mining and annotation process for constructing a human driving preference scenario dataset.} Only takeover cases related to driving styles are used to extract human preference labeling, which is updated by data annotation results.}
    \label{fig:fig4}
\label{fig4}
\end{figure}
We propose an efficient process for data mining and automatic annotation to build a dataset $\mathcal{D}_p$ that reflects human driving preferences, as shown in Fig.~\ref{fig:fig4}. This process involves extracting human preference definition from takeover data, mining scenarios according to rules or models, selecting preference key frame and conducting manual selection and model rejection.

\textbf{Preference Scenario Mining}
Human driving typically occurs in ordinary environments, making it difficult to define specific driving styles for every decision. For instance, choosing to overtake a slow-moving vehicle may not be classified as either aggressive or conservative. However, in critical scenarios—such as overtaking vehicles using the opposite lane, shown in Fig.~\ref{fig1}—distinct driving behaviors can be identified. These critical situations, referred to as preference scenarios, are mined from large-scale human driver takeover data. The data is categorized into six classes (e.g., ``too aggressive'' or ``too conservative''), each corresponding to a distinct driving style, which can be used to define rules or train models for identifying preference scenarios.

\textbf{Key Frame Selection}
After identifying preference scenarios, only segments relevant to the preference need to be labeled. Rather than labeling every frame, we focus on key frames where significant actions occur, such as changes in speed or direction. If a frame is tagged too early, the defining action has not yet happened; if tagged too late, the action is already underway. Clear specifications for key frame identification allow for rule-based, automatic detection, enabling the potential large-scale annotation.

\textbf{Post-verification}
The annotated key frames undergo random manual checks to ensure data quality. Human inspectors can update scenario definitions or introduce new preference scenarios in special cases. Model rejection uses baseline generative models to filter out the common cases that ground truth trajectories cover common average human preference knowledge. 

\subsection{Constructing semi-synthetic dataset for reward model training}
\label{app:ae}
 
 Utilizing preference dataset $\mathcal{D}_p=\{(o^{(j)};\mathbf{x}_c^{(j)})\}^n_{j=1}$ acquired by Appendix~\ref{app:datacollect}, we adopt a semi-synthetic approach that enables generating multiple trajectory preference pairs within the same scenario $o$. We use the pretrained model to predict trajectories $\mathbf{x}_{raw}$ for each scenario, randomly selecting $q$ trajectories to form $q$ raw trajectory preference pairs $\{(\mathbf{x}_c^{(j)},\mathbf{x}_{raw}^{(j)})\}^q_{j=1}$. Applied on whole  dataset $\mathcal{D}_p$, we acquires raw preference-pair dataset $\mathcal{D}_{raw}=\{(o^{(j)};\mathbf{x}_c^{(j)},\mathbf{x}_{raw}^{(j)})\}^{m}_{j=1}$, where $m=q*n$.

 However, trajectories directly generated by the model exhibit significant statistical differences from human-annotated trajectories, potentially leading to reward hacking during reward model training. To address this challenge, we present a trajectory reconstruction autoencoder $f_\eta$ that transfers the raw trajectory $\mathbf{x}_{raw}$ generated by our pre-trained model to a normalized trajectory $\mathbf{x}_i=f_\eta (\mathbf{x}_\text{raw})$. The architecture of the model is a symmetric 1D convolutional autoencoder with two-layer encoder and decoder blocks using ReLU activations. The model utilizes a dual loss function:
\begin{align}
\label{eq:loss_total}
\mathcal{L}^\eta
  &= (1-\delta)\,\mathcal{L}_{\text{recon}}
     +\delta\,\mathcal{L}_{\text{style}}, \\
\text{where }  \mathcal{L}_{\text{recon}}
  &= \lVert \mathbf{x}_i-\mathbf{x}_{raw} \rVert_2^{2}, \nonumber\\
\mathcal{L}_{\text{style}}
  &= \lVert \mathrm{mean}(\mathbf{v}_i)-\mathrm{mean}(\mathbf{v}_c) \rVert_2^{2}
   + \lVert \mathrm{std}(\mathbf{v}_i)-\mathrm{std}(\mathbf{v}_c) \rVert_2^{2}. \nonumber
\end{align}
The reconstruction loss $\mathcal{L}_{recon}$ maintains geometric consistency with the original diffusion-generated trajectories, and the style loss $\mathcal{L}_{style}$ transfers human-like motion characteristics by comparing velocity distribution statistics (means and variances) between generated and human-annotated trajectories. 

We train the autoencoder $f_\eta$ on a small subset of  $\mathcal{D}_{raw}$, apply $f_\eta$ to reconstruct $\mathbf{x}_{raw}$ and finally acquire semi-synthetic preference-pair dataset $\mathcal{D}_R=\{(o^{(j)};\mathbf{x}_c^{(j)},\mathbf{x}_i^{(j)})\}^{m}_{j=1}$.

\subsection{RL Modeling Details}\label{app:rl}

For a target driving style we collect a dataset
\(
\mathcal D_p=\{(o,\mathbf x_c)\}
\)
distributed as \(p(o,\mathbf x_c)\),
where \(o\) is a observation and
\(\mathbf x_c\) is a ground-truth trajectory.
Given \(o\), we draw a latent noise vector
\(z\sim\mathcal N(\mathbf0,\mathbf I)\)
and let the diffusion-policy
\(\pi_\theta\)
generate a candidate terminal trajectory
\(
\mathbf x_0\sim\pi_\theta(\mathbf x_0\mid o,z).
\)
Its scalar reward is then evaluated by
\(r_\phi(o,\mathbf x_0)\).

\paragraph{Mixed sampling distribution \(\mathcal P_\theta\).}
Denote a length-\(T\) denoising trajectory by
\(
\tau=(\psi_0,a_0,\psi_1,a_1,\dots,\psi_{T-1},a_{T-1},\psi_T),
\)
where the state–action pairs are defined as
\(\psi_t=(o,\mathbf x_{T-t})\) and
\(a_t=\mathbf x_{T-t-1}\).
The joint distribution that produces all randomness
in one rollout is

\begin{equation}\label{eq:mix_measure}
  \mathcal P_\theta(\tau)
  \;=\;
  p(o)\;
  \mathcal N(z)\;
  \prod_{t=0}^{T-1}
  \pi_\theta\bigl(a_t\mid\psi_t\bigr)\;
  \delta\!\bigl[\psi_{t+1}=(o,a_t)\bigr],
\end{equation}

where \(\delta[\cdot]\) is the Dirac measure that enforces the
deterministic state update, and
\(\mathbf x_0\) is obtained after the final denoising step
\(\psi_{T-1}\to a_{T-1}\).

\paragraph{Optimization objective.}
Using the mixed measure Eq.~\ref{eq:mix_measure}, the policy is trained to maximize

\begin{equation}\label{eq:app_objective}
J(\theta)
=\;
\mathbb E_{\tau\sim\mathcal P_\theta}
\bigl[
     r_\phi\bigl(o,\mathbf x_0\bigr)
\bigr].
\end{equation}

\vspace{4pt}
\noindent\textbf{MDP expansion.}
Following \citet{DDPO,DPPO}, the denoising process can be viewed as
a $T$-step MDP

\begin{equation}\label{eq:MDP}
\left\{
\begin{aligned}
\psi_t &\triangleq (o,\mathbf x_{T-t}), \\[2pt]
a_t &\triangleq \mathbf x_{T-t-1}, \\[2pt]
P(\psi_{t+1}\mid\psi_t,a_t) &\triangleq \delta_{(o,a_t)}, \\[2pt]
\pi_\theta(a_t\mid\psi_t) &\triangleq
      p_\theta(\mathbf x_{T-t-1}\mid\mathbf x_{T-t},o), \\[2pt]
\rho_0(\psi_0) &\triangleq p(o)\,\mathcal N(\mathbf0,\mathbf I), \\[2pt]
R(\psi_t,a_t) &\triangleq
  \begin{cases}
        r_\phi(o,\mathbf x_0), & t=T-1,\\
        0, & \text{otherwise}.
  \end{cases}
\end{aligned}
\right.
\end{equation}

Let
\(r_t(\psi_t,a_t)=\sum_{\tau=t}^{T-1}\gamma^{T-1-\tau}R(\psi_\tau,a_\tau)\)
be the discounted return from step \(t\).
The REINFORCE loss is then

\begin{equation}\label{eq:rlloss_reinforce}
\mathcal L_{\mathrm{RL}}(\theta)
=
-\,
\mathbb E_{\tau\sim\mathcal P_\theta}
\Bigl[
      \sum_{t=0}^{T-1}
      \log\pi_\theta(a_t\mid\psi_t)\,
      r_t(\psi_t,a_t)
\Bigr].
\end{equation}

\paragraph{Variance reduction.}
Instead of raw returns, we replace
\(r_t(\psi_t,a_t)\)
in Eq.~\ref{eq:rlloss_reinforce} with the
advantage estimator

\begin{equation}
    \hat A^{\pi_\theta}(\psi_t,a_t)
=
\gamma^{T-1-t}
\Bigl(
     r_\phi(o,\mathbf x_0)
     -\hat V^{\pi_\theta}(\psi_{T-1})
\Bigr),
\end{equation}

where
\(\hat V^{\pi_\theta}\)
is a learned critic.
To avoid critic warm-up and extra hyperparameters,
we subsequently adopt the GRPO method of
\citet{deepseek},
which normalizes terminal rewards within each
\(K\)-trajectory group and uses those estimated
group-relative advantages
$\hat{A}_{\mathrm{gr}}^{(k)}$ in place of
\(\hat A^{\pi_\theta}\), as shown in Eq.~\ref{gradv}. And
according to the reward definition and GRPO method, we simplified the REINFORCE loss Eq.~\ref{eq:rlloss_reinforce} to Eq.~\ref{rlloss}.

\section{Details of Experiments}
\subsection{EM Algorithm for Trajectory Optimization}
\label{app:EM}
Inspired by ~\citet{DBLP:journals/corr/abs-2111-14973}, we aim to find an optimal planning trajectory to represent the whole sampling trajectory distribution $\Psi$. This is achieved using an iterative clustering algorithm that functions similarly to Expectation-Maximization (EM) \citep{bishop2006pattern}, but with a hard assignment of cluster membership. 

First, we randomly pick one trajectory generation sample as the cluster centroid, denoted as $\bar{\mu}$. The objective is to maximize the probability that a centroid sampled from $\Psi$ is located within a distance $d$ of at least one of the selected centroids. This selection criterion, which explicitly promotes trajectory distribution representation, is formulated as:
\begin{equation}
\bar{\mu} = \underset{\bar{\mu}}{\mathrm{argmax}} \sum_{k=1}^{K} q_k \max_{\bar{\mu}} \mathbb{I}(\|\mu_k - \bar{\mu}\|_2 \leq d)
\end{equation}
    
where $q_k$ is the probability of sample $k$ from $K$ inferences and $\mathbb{I}(\cdot)$ is the indicator function.

Starting with the selected centroids, we iteratively update the parameters of $\bar{\Psi}$ using an Expectation-Maximization-style algorithm. Each iteration consists of the following updates:
\begin{align}
    \bar{q} &\leftarrow \sum_{k=1}^{K} q_k p(\mu_k; \bar{\Psi}) \\
    \bar{\mu} &\leftarrow \frac{1}{\bar{q}} \sum_{k=1}^{K} q_k p(\mu_k; \bar{\Psi}) x  \\
    \bar{\Sigma} &\leftarrow \frac{1}{\bar{q}} \sum_{k=1}^{K} q_k p(\mu_k; \bar{\Psi}) [\Sigma_k + (\mu_k - \bar{\mu})(\mu_k - \bar{\mu})^T]
\end{align}
Here, $p(x; \bar{\Psi})$ represents the probability that a sample $x$ is drawn from the model distribution specified by $\bar{\Psi}$.

\subsection{Human Preference BOE}
\label{app:boe}

Evaluating driving style requires human-level semantic understanding of the current scene. For example, when there is a cyclist in front of the ego vehicle, considering the context, we can determine that overtaking the cyclist is a more aggressive behavior compared to following behind the cyclist. When using overtaking behavior as the primary criterion for aggressive style, we found that ADE and FDE metrics do not effectively reflect this behavior. Compared to an aggressive ground truth trajectory, if a model's trajectory is overall faster or slower longitudinally, or has lateral deviations, these would result in larger ADE and FDE values. However, these longitudinal and lateral deviations are not clearly associated with whether the model generates an overtaking trajectory. Therefore, we choose to introduce human preference judgments to evaluate trajectory output styles. Specifically, based on their subjective assessment, evaluators select which trajectory better meets style requirements from a pair of trajectories in the same context.

Based on this approach, we designed a systematic human evaluation framework to compare our models. To evaluate the performance of models $m_1$ and $m_2$ on a set of scenes $\mathcal{D}$, we conduct human evaluation experiments. For each scene $o \in \mathcal{D}$, the two models generate trajectories $\mathcal{T}_{m_1}(o)$ and $\mathcal{T}_{m_2}(o)$ respectively. Evaluators are asked to choose from three options without knowing the source of the trajectories (double-blind assessment): prefer the trajectory generated by $m_1$, prefer the trajectory generated by $m_2$, or consider both equally effective in meeting the specified driving style.

Formally, we define the human evaluation function $h(\mathcal{T}_{m_1}(o), \mathcal{T}_{m_2}(o))$, whose value is:
\begin{itemize}
    \item $1$  if the evaluator prefers $\mathcal{T}_{m_1}(o)$
\end{itemize}
\begin{itemize}
    \item  $-1$ if the evaluator prefers $\mathcal{T}_{m_2}(o)$
\end{itemize}
\begin{itemize}
    \item $0$  if the evaluator considers both equally effective (tie)
\end{itemize}

For each scene $s \in \mathcal{S}$, we collected evaluations from $N$ evaluators.  Based on these evaluations, we define the  metric \textbf{Better Or Equal Rate (BOE)}: The proportion where one model is considered better than or equal to the other model

\begin{equation}
\left\{ \begin{aligned}
\text{BOE}_{m_1} = \frac{|\{o \in \mathcal{D} : h(\mathcal{T}_{m_1}(o), \mathcal{T}_{m_2}(o)) \geq 0\}|}{|\mathcal{D}|} \\
\text{BOE}_{m_2} = \frac{|\{o \in \mathcal{D} : h(\mathcal{T}_{m_1}(o), \mathcal{T}_{m_2}(o)) \leq 0\}|}{|\mathcal{D}|} \\
\end{aligned} \right.
\end{equation}

For the ``aggressive'' evaluation dataset, we manually select 53 clips from the test set where the pretrained model's performance exhibit significant deviations from the expected driving style. For the ``defensive'' evaluation dataset, we manually select 151 clips with similar characteristics. We recruit 5 professional annotators who regularly perform driving style labeling tasks to serve as evaluators. These evaluators are instructed to make preference judgments on trajectory outputs based on the specified driving style requirements and scene context. To mitigate individual bias, the preference metrics for each model are derived by averaging the scores across all evaluators, ensuring a more objective assessment of the models' performance.

\subsection{Internal Data Capacities}
\label{app:data}
The table contains the capacity of each split in Internal dataset, depicted by scenario clips and frames.
\begin{table}[htbp]
    \renewcommand\arraystretch{1.2}
    \centering
    \caption{\textbf{Dataset capacities of different splits of Internal dataset.} Small, medium and large splits refer to Internal Normal dataset. ``Aggressive'' and ``defensive'' splits are from Internal Preference dataset.}
    \begin{tabular}{lccccc}
        \hline
        \textbf{Data Split}  & Clips & Frames \\     
        \hline
        small & 65,628 & 810K  \\
        medium & 298,706          & 3M \\
        large & 487,812          & 6M  \\
        aggressive & 2967          & 35329 \\
        defensive  & 49030    & 136782 \\
        \hline \\
    \end{tabular}
    \label{table:data}
\end{table}

\subsection{Implementation Details}
\label{app:imp}
Our model employs a base ViT ~\citep{dosovitskiy2020image} as the image backbone and ResNet34 ~\citep{he2016deep} as the LiDAR backbone. The input front-view images are concatenated at a resolution of $1024\times 256$, while the LiDAR is $256\times 256$. We adopt an SDE-based denoising paradigm with 10 steps for both training and inference. The learning rate is set to be $1e-4$, and the batch size is $512$, distributed across $8$ NVIDIA-H20-RDMA GPUs. We use the $OneCycle$ scheduler and the Adam optimizer with default parameters. As required by the task, the model outputs $8$ waypoints spanning $4$ seconds. The model is trained for $1000$ epochs as the base model on NavSim benchmark. 

In the semi-synthetic dataset construction (Appendix~\ref{app:ae}), the sampling rate $q=3$. For the trajectory reconstruction model training, the weight of style $\delta=0.3$, the learning rate is $1e-3$, the batch size is $32$ and the epoch number is $100$. In reward model training, we adopt margin constant $m=1$, the learning rate is $2e-5$, the batch size is $32$ and the max epoch number is $30$ with early-stop mechanism.

For NavSim benchmark, we finetune the base model on NavSim trainval split using GRPO~\citep{deepseek} and employ the PDMS metric as the reward. The pretrained model is fine-tuned for $45$ epochs using $16$ NVIDIA-H20-RDMA GPUs with a per-GPU batch size of $16$, a BC loss weight of $1e-2$, a learning rate of $6e-5$ for the ViT component, and $5e-5$ for the remaining parts. The EM algorithm iterates 25 times for each scenario.

For supervised learning (SL) and reinforcement learning (RL) experiment, we set sample number $K=8$ and discount factor $\gamma =0.99$. Behavior cloning loss weights $\alpha$ is set to $1e-1$ in Section~\ref{sec:rlft} and varies from $1$ to $1e-3$ in Appendix~\ref{app:bc}. The learning rate is $5e-5$  and the batch size is $128$ for the SL and $32$ for the RL. We adopt $200$ epoch training for SL and $20$ epoch training for RL. The short supervised refresh takes the same setting as SL and is trained for $100$ epoch.

\subsection{Qualitative Analysis}
\label{app:qual}
In Fig.~\ref{fig:compare}, \systemname\ achieves excellent performance qualitatively for both ``aggressive'' and ``defensive'' styles.

For ``aggressive'' finetuning, in the first case, there are Vulnerable Road Users (VRU) and a parked vehicle ahead in the ego lane, and the ego vehicle is about to pass a vehicle in the oncoming left lane, representing a scenario with strong interactions. The pretrained policy plans trajectories that tend to follow behind the VRUs, showing a conservative behavior. However, after RL finetuning, the policy plans a trajectory that borrows the opposite lane to bypass the VRUs, becoming more aggressive and efficient.
The second case is similar to the first one, where the ego vehicle intends to bypass a VRU. Although the pretrained policy also plans lane-changing trajectories to bypass, their speeds are not fast enough. After RL finetuning, the policy significantly increases the driving speed.
In the third case, the pretrained strategy plans trajectories following the slow-moving vehicle ahead in the current lane, exhibiting inefficient behavior. After RL finetuning, the planned trajectories involve a left lane change to overtake the preceding vehicle, thereby improving driving efficiency, making it more consistent with the aggressive human driving ground truth trajectory.

The effectiveness of finetuning with defensive driving preferences is evident in several qualitative cases. In the first and second scenarios, the finetuned model exhibits cautious behavior by proactively decelerating in potentially ambiguous or high-risk situations, such as approaching an intersection or navigating under low-visibility nighttime conditions. These adaptations reflect a heightened awareness of environmental complexity, which is not observed in the baseline model that maintains a higher velocity without accounting for contextual risk.
In the third scenario, where a large vehicle initiates an overtaking maneuver, the defensive-style planner successfully generates a trajectory that yields space and reduces velocity to mitigate potential conflict. In contrast, the original model, lacking an explicit preference for conservative behavior, fails to adjust its trajectory and continues at its nominal speed, increasing the risk of lateral interaction. These results demonstrate that the preference-aligned model not only respects safety constraints but also anticipates socially appropriate responses to multi-agent interactions, contributing to more robust and human-aligned motion planning.

\section{Supplementary Ablation Experiments}
\label{app:sup}

\subsection{Ablation for PPO Variants}
\label{app:ppov}
Preference fine-tuning for diffusion-based trajectory prediction is a highly novel and underexplored task. The most closely related prior work is \citet{wang2024reinforcement}. Since their reward model, dataset, and codebase are not publicly available, we re-implement the PPO-based preference optimization procedure described in their paper for comparison in our ablation study. The results on both our preference test set and the standard test set are summarized in the table below.

\begin{table}[htbp]
    \renewcommand\arraystretch{1.2}
    \centering
    \caption{Results of \citet{wang2024reinforcement} Method}
    \begin{tabular}{llccccc}
        \hline
        \textbf{Method} & \textbf{Test Split} & \textbf{minADE} & \textbf{meanADE} & \textbf{minFDE} & \textbf{meanFDE} & \textbf{Diversity} \\     
        \hline
        Aggr. PPO & Aggressive & 0.7099 & 1.5640 & 1.7958 & 4.2434 & 0.4704 \\
        Aggr. PPO & Normal     & 0.2716 & 0.8546 & 0.6518 & 2.3182 & 0.3397 \\
        \hline
        Defensive PPO & Defensive & 0.4638 & 1.3630 & 1.2667 & 4.1038 & 0.3754 \\
        Defensive PPO & Normal    & 0.3314 & 1.1452 & 0.8002 & 3.1791 & 0.3346 \\
        \hline
    \end{tabular}
    \label{tab:PPO_Variant}
\end{table}

The results indicate that while the re-implemented PPO algorithm does work for the task, its performance on the preference test set is inferior to our DPGRPO-finetuned model. Its performance on the standard (normal) test set suggests that the PPO-tuned model remains closer to the original pretraining distribution, implying limited adaptation to preference signals.

We conjecture PPO underperforms DPGRPO mainly for two reasons. First, the regularization and clipping in the PPO variant from \citet{wang2024reinforcement}, while effective for their simpler lane-change task, are too restrictive for our complex multi-scenario setting. These constraints risk collapsing the policy into a narrow mode, failing to capture the multimodal distributions that diffusion models can generate and that diverse driving contexts demand. In contrast, GRPO's group-advantage mechanism (Fig.~\ref{fig:fig2}) aligns well with the multimodal nature of the diffusion generator, thereby better unlocking its potential in our preference-tuning framework.

Furthermore, we evaluate Direct Preference Optimization (DPO)\citep{rafailov2023direct} as an alternative preference‑fine‑tuning approach; the corresponding results are shown below:

\begin{table}[htbp]
    \renewcommand\arraystretch{1.2}
    \centering
    \caption{Results of DPO-finetuned Model}
    \begin{tabular}{llccccc}
        \hline
        \textbf{Method} & \textbf{Test Split} & \textbf{minADE} & \textbf{meanADE} & \textbf{minFDE} & \textbf{meanFDE} & \textbf{Diversity} \\     
        \hline
        Aggr. DPO & Aggressive & 4.9720 & 5.1282 & 10.2502 & 10.7709 & 0.1835 \\
        \hline
    \end{tabular}
    \label{tab:DPO-finetuned}
\end{table}

We find DPO to be ineffective for preference fine-tuning in autonomous driving, as it struggles with the task's complex, multi-objective nature. DPO tends to overfit to a single dimension of the preference signal instead of balancing competing objectives, which often leads to training instability. This issue is compounded by DPO's high sensitivity to the noisy and ambiguous labels inherent in preference data collected from diverse, real-world takeover scenarios.

\subsection{Ablation for BC Loss Weight}
\label{app:bc}
This table shows the ablation results for varying behavior cloning (BC) loss weights $\alpha$, which is tested on the Internal preference test split.
\begin{table}[htbp]
    \renewcommand\arraystretch{1.2}
    \centering
    \caption{Impact of Behavior Cloning Loss Weight.}
    \begin{tabular}{lcccccc}
        \hline
        \textbf{$\alpha$} & Data Split   & minADE$\downarrow$ & meanADE$\downarrow$ & minFDE$\downarrow$ & meanFDE$\downarrow$ & Diversity$\uparrow$ \\     
        \hline
        1.0 & Aggressive& 0.5535 & 1.5238          & 1.3422 & 4.1555         & 0.6035 \\
        0.5 & Aggressive& 0.5635          & 1.5747          & 1.3694         & 4.2782          & 0.6088 \\
        1e-1 & Aggressive& \textbf{0.5304}          & \textbf{1.5036} & \textbf{1.2788}          & \textbf{4.1039}          & 0.6051 \\
        1e-2 & Aggressive& 0.5585          & 1.5697          & 1.3479          & 4.2692 & \textbf{0.6142} \\
        1e-3   & Aggressive & 0.5330          & 1.5332          & 1.2805         & 4.1650          & 0.6131 \\
        \hline
        1.0 & Defensive& 0.3187 & 1.5219          & 0.8418 & 4.5173          & 0.3763 \\
        0.5 & Defensive& 0.3252          & \textbf{1.4671}          & 0.8579          & \textbf{4.3304}          & 0.3758 \\
        1e-1 & Defensive& \textbf{0.3101}          & 1.5309 & \textbf{0.8092}          & 4.5603         & \textbf{0.3900} \\
        1e-2 & Defensive& 0.3209          & 1.4827          & 0.8389          & 4.3797 & 0.3755 \\
        1e-3   & Defensive & 0.3195          & 1.5972          & 0.8368          & 4.6734          & 0.3782 \\
        \hline \\
    \end{tabular}
    \label{table:bc}
\end{table}

The experimental results across multiple datasets demonstrate that  $\alpha=1e-1$ consistently achieves optimal performance for both preference types. This optimal $\alpha$ balances exploration in preference space while preventing the policy from diverging too far from reasonable driving behaviors.

\subsection{Ablation of data scales of DPGRPO}
\label{app:datascale}
We add a data‑efficiency ablation study. The results obtained with 10\% and 50\% random subsamples of the preference data are as follows:

\begin{table}[htbp]
    \setlength{\tabcolsep}{3pt}
    \renewcommand\arraystretch{1.2}
    \centering
        \caption{Impact of data scales of DPGRPO. ``Aggr.'' and ``Defen.'' refer to the aggressive and defensive datasets used for training and evaluation.}
    \begin{tabular}{lllccccc}
        \hline
        \textbf{Method} & \textbf{Scale} & \textbf{Test Split} & \textbf{minADE}$\downarrow$ & \textbf{meanADE}$\downarrow$ & \textbf{minFDE}$\downarrow$ & \textbf{meanFDE}$\downarrow$ & \textbf{Diversity}$\uparrow$ \\
        \hline
        Aggr. RL & 10\% & Aggr. & 0.7025 & 2.1477 & 1.6146 & 5.5560 & \textbf{0.6437} \\
        Aggr. RL & 50\% & Aggr. & 0.5431 & 1.5623 & 1.3091 & 4.2650 & 0.6015 \\
        Aggr. RL & 100\% & Aggr. & \textbf{0.5304} & \textbf{1.5036} & \textbf{1.2788} & \textbf{4.1039} & 0.6051 \\
        \hline
        Defen. RL & 10\% & Defen. & 0.4483 & 1.8836 & 1.1182 & 5.2600 & 0.3746 \\
        Defen. RL & 50\% & Defen. & 0.3352 & 1.5785 & 0.8638 & 4.6677 & \textbf{0.4014} \\
        Defen. RL & 100\% & Defen. & \textbf{0.3101} & \textbf{1.5309} & \textbf{0.8092} & \textbf{4.5603} & 0.3900 \\
        \hline
    \end{tabular}
    \label{tab:data_scales}
\end{table}

The experiments reveal a logarithmic saturation trend: performance improves substantially when the data budget increases from 10\% to 50\%, whereas the marginal gain from 50\% to 100\% is much smaller. This aligns with the normal data‑efficiency behavior in reinforcement learning and underlines the effectiveness of the proposed DPGRPO algorithm.

\end{document}